\documentclass[10pt,twocolumn,letterpaper]{article}

\usepackage[pagenumbers]{cvpr} 

\usepackage{graphicx}
\usepackage{amsmath}
\usepackage{amssymb}
\usepackage{booktabs}
\usepackage{bm}
\usepackage{multirow}
\usepackage{caption,subcaption}
\usepackage{verbatim} 
\usepackage{balance}

\usepackage[pagebackref,breaklinks,colorlinks]{hyperref}

\usepackage[capitalize]{cleveref}
\crefname{section}{Sec.}{Secs.}
\Crefname{section}{Section}{Sections}
\Crefname{table}{Table}{Tables}
\crefname{table}{Tab.}{Tabs.}

\def\cvprPaperID{10675} 
\def\confName{CVPR}
\def\confYear{2023}

\begin{document}

\title{Multi-video Moment Ranking with Multimodal Clue}

\author{Danyang Hou$^{1, 2}$, Liang Pang$^{1}$, Yanyan Lan$^4$, Huawei Shen$^{1,3}$, Xueqi Cheng$^{2,3}$\\
$^1$ Data Intelligence System Research Center, Institute of Computing Technology, CAS, Beijing, China \\
$^2$ CAS Key Lab of Network Data Science and Technology, \\ Institute of Computing Technology, CAS, Beijing, China\\
$^3$ University of Chinese Academy of Sciences, Beijing, China \\
$^4$ Institute for AI Industry Research, Tsinghua University, Beijing, China \\
}
\maketitle

\begin{abstract}
Video corpus moment retrieval~(VCMR) is the task of retrieving a relevant video moment from a large corpus of untrimmed videos via a natural language query. 
State-of-the-art work for VCMR is based on two-stage method.
In this paper, we focus on improving two problems of two-stage method: (1)~Moment prediction bias: The predicted moments for most queries come from the top retrieved videos, ignoring the possibility that the target moment is in the bottom retrieved videos, which is caused by the inconsistency of Shared Normalization during training and inference. (2)~Latent key content: Different modalities of video have different key information for moment localization. To this end, we propose a two-stage model \textbf{M}ult\textbf{I}-video ra\textbf{N}king with m\textbf{U}l\textbf{T}imodal clu\textbf{E}~(MINUTE). MINUTE uses Shared Normalization during both training and inference to rank candidate moments from multiple videos to solve moment predict bias, making it more efficient to predict target moment. In addition, Mutilmdaol Clue Mining~(MCM) of MINUTE can discover key content of different modalities in video to localize moment more accurately. MINUTE outperforms the baselines on TVR and DiDeMo datasets, achieving a new state-of-the-art of VCMR. Our code will be available at GitHub.

\end{abstract}

\section{Introduction}
\label{sec:intro}

\begin{figure}
    \begin{center}
    \includegraphics[width=0.35\textwidth]{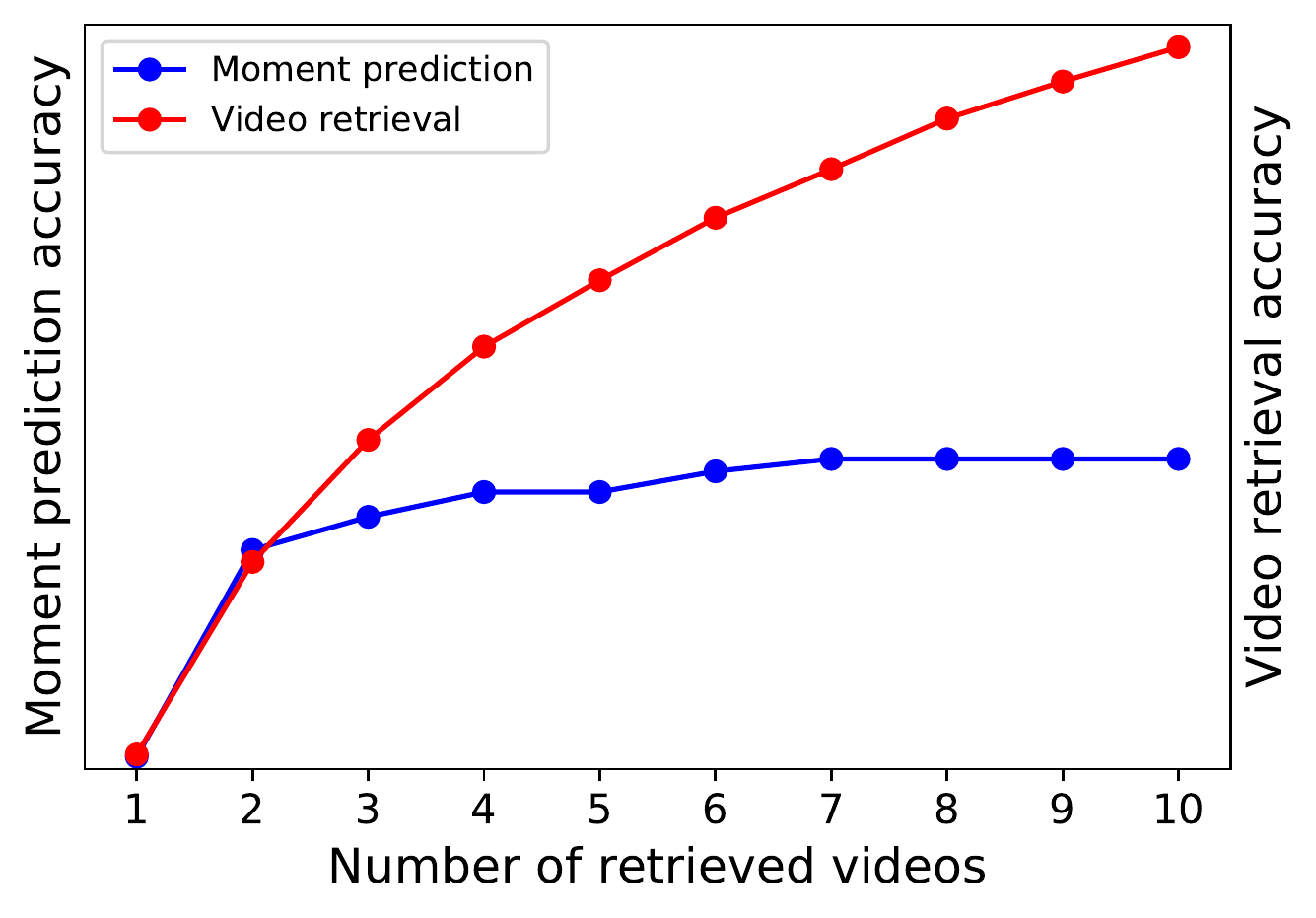}
    \end{center}
    \caption{\textbf{Moment prediction bias}: Video retrieval accuracy improves as the number of retrieved videos increases, indicating that the probability of predicting the correct moment also increases. However, when the number of retrieved videos exceeds 2, moment prediction accuracy hardly increases, which means that predicted moments for most queries come from the top 2 videos.}
    \label{fig:bias}
\end{figure}

The rise of video-sharing applications has led to a dramatic increase in the number of videos on the Internet. Faced with such a huge video corpus, users need an accurate retrieval tool to meet the needs of fine-grained cross-modal information. We have the opportunity to address this challenge thanks to the recently proposed video corpus moment retrieval~(VCMR)~\cite{temporal, lei2020tvr} task that requires retrieving a video moment via a natural language query from a collection of untrimmed videos, where the moment is a temporal segment of a video. VCMR consists of two sub-tasks: video retrieval~(VR) and single video moment retrieval~(SVMR). The goal of VR is to retrieve videos that may contain the target moment via a natural language query. And SVMR aims to use the query to localize the target moment in the retrieved videos.

According to different strategies to learn two sub-tasks, existing methods can be divided into one-stage method and two-stage method. One-stage method~\cite{lei2020tvr, zhang2021video, li2020hero, zhang2020hierarchical} treats VCMR as a multi-task learning problem, using a shared backbone with two different heads to learn VR and SVMR. Whereas two-stage method~\cite{hou2021conquer} leverages a pipeline of two independent modules to learn the two sub-tasks. Specially, it first trains a video retriever by query-video pairs to learn VR, then takes advantage of Shared Normalization~(Shared-Norm)~\cite{clark-gardner-2018-simple} technique to train localizer to learn SVMR, where the negatives for Shared-Norm are from the training data sampled by the trained retriever. 
In inference, it first uses retriever to select the most relevant K videos from corpus, then uses localizer to localize the candidate moments in the K videos. The final predicted moment depends on both retrieval score and localization score. Two-stage method is more suitable for VCMR because ~(1) Shared-Norm can enhance the possibility of the target moment appearing in the correct video.~(2) Two-stage method can select models with different query-video interaction modes in the two modules. For example, it select late-fusion model as retriever for fast video retrieval, and leverage early-fusion model as localizer for accurate moment localization. State-of-the-art model~\cite{hou2021conquer} for VCMR is also based on two-stage method.

\begin{figure}
    \begin{center}
    \includegraphics[width=0.50\textwidth]{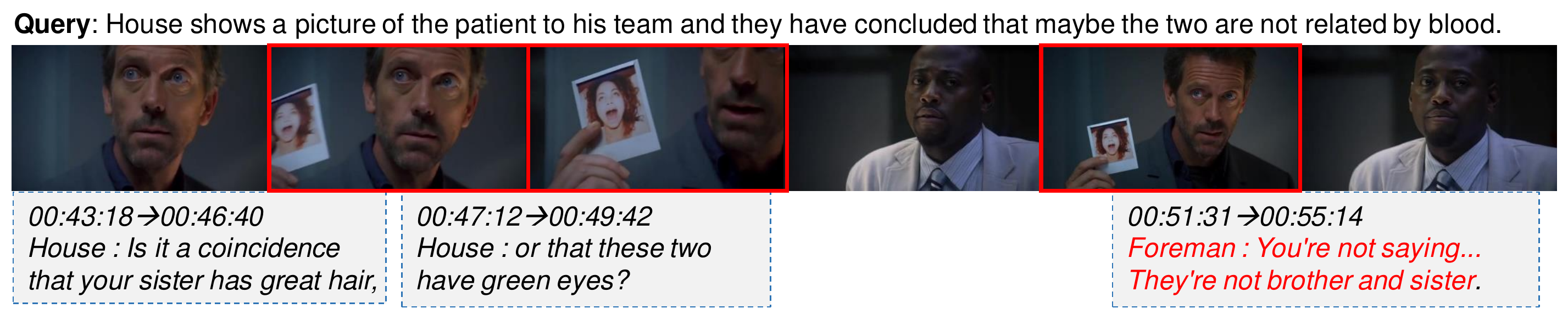}
    \end{center}
    \caption{\textbf{Latent key content}: The images with a red border are visual key content because these are relevant to “House shows a picture of the patient to his team” in query. The highlighted subtitle is textual key content, for it relates to "they have concluded that maybe the two are not related by blood".}
    \label{fig:clue}
\end{figure}

However, two problems limit the performance of two-stage method. The first is \textbf{Moment prediction bias}: as shown in \cref{fig:bias}, the final predicted moments for most queries are from the top-ranked videos among the K retrieved videos. This is counter-intuitive because the more videos retrieved, the more likely those videos contain the correct moment.
This bias neglects the possibility that the target moment is in the bottom-ranked videos.
The reason for this bias is that although two-stage method uses Shared-Norm to normalize the probability of correct moment across correct video and negative videos, it still only normalizes the probability of the candidate moments in a single video during inference. This inconsistency in training and inference results in the incomparable localization scores of candidate moments during inference. Since the final predicted moment depends on both video retrieval score and moment localization score, the incomparable localization scores  will make the final moment mainly depend on video retrieval scores, resulting in the final predicted moment more tending to come from videos with higher rankings. The second problem is \textbf{Latent key content}: the localizer of two-stage method neglects key content from different modalities for moment localization. Video is usually composed of multimodal information, such as images~(vision) and subtitles~(text). As shown in \cref{fig:clue}, visual information and textual information have different emphases, if we can find out the important visual information and textual information as clues, it will help better moment localization. 

In this paper, we propose \textbf{M}ult\textbf{I}-video ra\textbf{N}king with m\textbf{U}l\textbf{T}imodal clu\textbf{E}~(MINUTE) to improve the two problems of two-stage method. For the first problem, we keep the consistence of Shared-Norm between training and inference, which forces the localization scores of candidate moments among multiple videos retrieved by retriever to be comparable during inference. 
On this basis, we derive a new scoring function to rank the candidate moments, which can combine the scores of video retrieval and moment localization more effectively. For the second problem, we propose an early-fusion localizer with a Multimodal Clue Mining~(MCM) component which can discover key content from different modalities to help moment localization. Specially, MCM first uses query to measure the importance of all images and subtitles in the video, then assigns weights to these elements according to their importance. The elements with high importance can be seen as key clues to improve moment localization. Then we feed weighted video representation together with query representation to a multimodal Transformer that captures deeper interactions between video and query to predict moments.

We conduct extensive experiments on TVR and DiDeMo datasets. The experimental results show that our proposed MINUTE outperforms other baselines, achieving a new state-of-the-art result. Ablation experiments verify that our method improves the two problems of two-stage method.

\section{Related Work}
 We first briefly introduce works related to two sub-tasks of VCMR. After that, we introduce recent works for VCMR in detail.

\noindent \textbf{Text-video retrieval} is a cross-modal retrieval task whose goal is to retrieve relevant videos from a corpus through a natural language query. This task is similar to VR of  VCMR task, but most content of the video in the former is relevant to the query, while only a small part of the content of the video in the latter is relevant to the query. The works for text-video retrieval can be divided into two categories depending on the interaction mode between query and video, e.g., late fusion and early fusion. Late-fusion methods~\cite{dong2021dual, patrick2020support, yang2020tree}  use two separated encoders to embed images and videos into a shared semantic space.
These models can be very efficient if we calculate and index each modal representation offline, for only the similarity between video and query should be applied in inference. 
Early-fusion methods~\cite{chen2020fine, gabeur2020multi, wang2021t2vlad} make fine-grained interactions between video and query with an attention mechanism~\cite{Bahdanau_attention, vaswani2017attention} to improve retrieval accuracy. 

\noindent \textbf{Temporal language grounding}  is a task similar to SVMR, which requires localizing a moment from a video given a natural language query.  Temporal language grounding can be seen as a special case of VCMR, with only one video in the corpus for each query.   According to the way of predicting moment, the existing works for temporal language grounding can be divided into proposal-based and proposal-free. 
Proposal-based method~\cite{gao2017tall,chen2019semantic,xiao-etal-2021-natural,chen2018temporally,zhang2021multi, liu2021context} first generates several proposals as candidates and then ranks the proposals according to their matching degree with the query, and the proposal with the highest matching degree is regarded as the answer. 
Unlike the proposal-based method, proposal-free method~\cite{yuan2019find,chen2020rethinking,zeng2020dense,li2021proposal,zhang2020span} directly predicts the start and end times of the moment without pre-extracting proposals as candidates.  

\noindent \textbf{Video corpus moment retrieval} is first proposed by~\cite{temporal}, then~\cite{lei2020tvr} propose a new dataset TVR for VCMR who extends the uni-modal video (image) in the previous dataset video to multiple modalities  (image and subtitle). 
The existing works for VCMR can be divided into two categories depending on how they learn the two sub-tasks, e,g., one-stage~\cite{lei2020tvr, zhang2021video, li2020hero, zhang2020hierarchical} method and two-stage method~\cite{hou2021conquer}. The one-stage method treats VCMR as a multi-task learning problem , using a shared model with two different heads to learn VR and SVMR simultaneously. XML~\cite{lei2020tvr} is the first one-stage method for VCMR who uses a late-fusion model to encode video and query separately and then uses two different heads to learn the two tasks. ReLoCLNet~\cite{zhang2021video} leverage contrastive learning to enhance the performance of XML.~\cite{li2020hero} also follows XML and proposes a video-language pre-train model HERO, which significantly improves the performance. HAMMER\cite{zhang2020hierarchical} is an early-fusion one-stage model that uses attention to make deep interactions between query and video for more accurate moment retrieval. 
Two-stage method leverages two different modules to learn two sub-tasks. CONQUER~\cite{hou2021conquer} is the only two-stage method that uses video retrieval heads of HERO~\cite{li2020hero} as the retriever and proposes a model based on context-query attention (CQA)~\cite{yu2018qanet} as the localizer. CONQUER achieves state-of-the-art results on VCMR. In training, CONQUER uses Shared-Norm~\cite{clark-gardner-2018-simple} technique to train localizer. In inference, CONQUER first uses a video retriever to retrieve top-K videos, then uses a moment localizer to localize the moment in the retrieved videos. Two-stage method is more suitable for VCMR, but it suffers from \textbf{moment prediction bias} and \textbf{latent key content}. In this paper, we focus on improving the two problems.

\section{Background}
We first formulate VCMR, then describe two-stage method, followed by analyzing moment prediction bias.

\subsection{Task Formulation}
We denote a corpus of videos $\mathcal{V} = \{v_1, v_2,...,v_{|\mathcal{V}|}\}$ where $|\mathcal{V}|$ is the number of videos in corpus and $v_i=\{f^1_i,f^2_i,...,f^{|v_i|}_i\}$ the $i$-th video which contains $|v_i|$ frames. Each frame $f_i^j$ consists of an image and a subtitle $(I^j_i, s^j_i)$. Note that if it contains no subtitle, $s^j_i$ is set to empty. 

Given a natural language query $q=\{w^1,w^2,...,w^{|w|}\}$ which consists of a sequence of words, the goal of VCMR is to retrieve most relevant moment $m_*$ from $\mathcal{V}$.  The target moment $m_*$ is a temporal segment $(\tau_{*,st}, \tau_{*, ed})$ in video $v_*$, where $v_*$ denotes the video that contains the target moment whose start and end timestamps are  $\tau_{*, st}$ and $\tau_{*, ed}$ respectively. 

The goal of VCMR can be seen as maximizing the probability of target moment $m_*$ given the query $q$ and the video corpus $\mathcal{V}$:
\begin{equation}
\label{equ:goal}
m_* = \mathop{\rm argmax}\limits_{m} P(m|q, \mathcal{V}).
\end{equation}
According to the chain rule of conditional probability:
\begin{equation}
\label{equ:two_tasks}
P(m_*|q, \mathcal{V}) = P(m_*|v_*,q) \cdot P(v_*|q, \mathcal{V}) ,
\end{equation}
where $P(v_*|q, \mathcal{V})$ and $P(m_*|v_*,q)$ are the probabilities of retrieving a video $v_*$ from corpus $\mathcal{V}$ and localizing the target moment $m_*$ in the retrieved video respectively.
The probability of target moment depends on the probabilities of start and end timestamps:
\begin{equation}
\label{equ:st_ed_prob}
P(m_*|v_*,q) = P_{st}(\tau_{*, st}|v_*,q)\cdot P_{ed}(\tau_{*, ed}|v_*,q). 
\end{equation}

\subsection{Two-stage Method}

Two-stage method uses a video retriever to model $P(v_*|q, \mathcal{V})$ and a moment localizer to model $P(m_*|v_*,q)$. 

In training, two-stage method use margin-based loss~\cite{faghri2017vse++} to train video retriever, then use Shared-Norm to train moment localizer. Specially, for a query, there is a positive video $v^+$ whose moment $(\tau_{+,j}, \tau_{+,k})$ is ground truth and $n$ negative videos $\{ v^-_1, v^-_2, \dots,  v^-_n\}$ that do not contain target moment. Shared-Norm is leveraged to normalize the probabilities of $\tau_{*,j}$ as start time and $\tau_{*,k}$ as end time  across all frames in positive video and negatives, such as:
\begin{equation}
\label{equ:train}
P_{st}(\tau_{+,j}|v^+,q) = \frac{{\rm exp}(l^{st}_{+,j})}{\sum \limits_{a=1} \limits^{n+1} \sum \limits_{b=1} \limits^{|v_b|} {\rm exp}(l^{st}_{a,b}) } ,
\end{equation}
where $l^{st}_{a,b}$ is the the logits that $b$-th frame in video $v_a$ is start timestamp of ground truth moment, and $|v_b|$ is the number of frame in a video. 
Training with Shared-Norm enhances the possibility of the target moment existing in the correct video. 

In inference, the retriever first uses the query to retrieve top-K videos from the corpus, then the localizer localizes the target moment in the retrieved videos. The score of the final predicted moment $(\tau_{i,j}, \tau_{i,k})$ in video $i$ with start time $j$ and end time $k$ depends on both retrieval score and localization score, the scoring function is:
\begin{equation}
\label{equ:score_pre}
S_{i,jk} = {\rm exp}(\alpha \cdot S^R_i) \cdot S^L_{i,jk} ,
\end{equation}
where $S_{i,jk}$ is the final score of the predicted moment, $S^R_i$ is the retrieval score of video $v_i$ , and $S^L_{i,jk}$ is the localization score of a moment in a video, and $\alpha$ is a hyper-parameter to encourage the target moment from top retrieved videos. The retrieval score is computed by cosine similarity between query representation and video representation. And the localization score is computed by the probability of a moment in a single video:
\begin{equation}
S^L_{i,jk} = P_{st}(\tau_{i,j}|v_i,q) \cdot P_{ed}(\tau_{i,k}|v_i,q) , 
\end{equation}
where $P_{st}(\tau_{i,j}|v_i,q)$ or $P_{ed}(\tau_{i,k}|v_i,q)$ is normalized across a single video : 
\begin{equation}
\label{equ:inference}
P_{st}(\tau_{i,j}|v_i,q) = \frac{{\rm exp}(l^{st}_{i,j})}{ \sum \limits_{b=1} \limits^{|v_i|} {\rm exp}(l^{st}_{i,b}) } ,
\end{equation}

\subsection{Moment Prediction Bias}
As shown in \cref{fig:bias}, the final predicted moments of two-stage method for most queries come from top-ranked videos. This bias limits the performance of two-stage method on VCMR, because it neglects the possibility of the target moment existing in the bottom-ranked videos.  We conjecture that this bias mainly comes from the inconsistency of normalization during training and inference, shown in \cref{equ:train} and \cref{equ:inference}. 

In training, it uses Shared-Norm to highlight the significance of the correct moment being in the correct video. Nevertheless, in inference, this probability is based on every single video, resulting in the predicted candidate moments from different videos being incomparable, so the significance no longer exists. Therefore, the score of the final predicted moment in \cref{equ:score_pre} is more dependent on video retrieval score, making the final predicted moment more likely to be from the top-ranked videos.

\section{Method}
We first illustrate how we improve moment prediction bias. Then we introduce the proposed model MINUTE, we emphasize multimodal clue mining component. Finally, we describe the training of MINUTE.

\subsection{Multi-video Moment Ranking in Prediction}
\label{sec:method_bias}
We propose to adopt Shared-Norm in inference, so that the localization scores of candidate moments from multiple videos are comparable, which can enhance the influence of moment localization score $S^L_{i,jk} $ on the final score $S_{i,jk}$ to improve moment prediction bias. Furthermore, we derive a new scoring function from \cref{equ:two_tasks} to combine the video retrieval and moment localization scores more effectively.

Specially, to compute $P(v_*|q, \mathcal{V})$, we obtain video representation $\bm{v}_i=\{\bm{f}^1_i,\bm{f}^2_i,...,\bm{f}^{|v_i|}_i\}$ and query representation $\bm{q}$. In the following paper, we use bold notations to denote vectors. The $j$-th frame representation $\bm{f}^j_i$ consists of image representation and subtitle representation $(\bm{I}^j_i, \bm{s}^j_i)$. Query also has two representations $(\bm{q}^I, \bm{q}^s)$ to compute similarity scores for images and subtitles respectively. The query and video representations details are in \cref{sec:retrieval}. 

Because only part of the content in the video is related to the query, the similarity score between the query and video $S^R_i$ is the average of max-pooling of query-image scores and max-pooling of query-subtitle scores. We use the inner product as the similarity score ${\rm sim}()$:
\begin{equation}
\begin{aligned}
{\rm sim}(\bm{q^c},\bm{c}_i^j) &= {\bm{q^c}}^T \cdot \bm{c}_i^j, \ \ c \in \{I,s\},  \\ 
\phi_c  &= \mathop{\rm max}\limits_{1 \leq j \leq |v_i|} {\rm sim}(\bm{q^c},\bm{c}_i^j),  \\
S^R_i &= \frac{\phi_I + \phi_s}{2}.
\end{aligned}
\end{equation}

The probability  $P(v_*|q, \mathcal{V})$ is computed by softmax normalized score across all query-video scores in corpus:
\begin{equation}
\label{equ:prob_retrieval}
P(v_* | q, \mathcal{V}) = \frac{{\rm exp}(S^R_*)}{\sum_{j=1}^{|\mathcal{V}|} {\rm exp}(S^R_j)}.
\end{equation}
Computing the inner product between query and all videos in the corpus is computationally intensive, so we employ Max Inner Product Search (MIPS)~\cite{2014Asymmetric} to find top-K videos to approximate the probability. The calculation of $P(v_* | q, \mathcal{V})$ in \cref{equ:prob_retrieval} can be approximated by $P(v_* | q, \mathcal{V}_*)$:
\begin{equation}
\label{equ:prob_topk}
P(v_* | q, \mathcal{V}) \approx P(v_* | q, \mathcal{V}_*) =  \frac{{\rm exp}(S^R_*)}{\sum_{j=1}^{K} {\rm exp}(S^R_j)}.
\end{equation}
The probabilities of the rest videos in the corpus are considered close to 0. The training of the retriever is to maximize the log-likelihood of probability ${\rm log} P(v_* | q, \mathcal{V})$, which is different from the previous two-stage method who use margin-based loss.

As for $P(m_*|v_*,q)$, we use Shared-Norm in inference, which is consistent with that in training to improve moment prediction bias:
\begin{equation}
\label{equ:prob_localization}
\begin{split}
P(m_*|v_*,q) & \approx  P(m_*|\mathcal{V}_*,q) = \\
&\frac{{\rm exp}(l^{st}_{*,j})}{ \sum \limits_{a=1} \limits^{K} \sum \limits_{b=1} \limits^{|v_i|} {\rm exp}(l^{st}_{a,b}) } \cdot \frac{{\rm exp}(l^{ed}_{*,k})}{\sum \limits_{a=1} \limits^{K}  \sum \limits_{b=1} \limits^{|v_i|} {\rm exp}(l^{ed}_{a,b}) }.
\end{split}
\end{equation}
A well-trained localizer should suppress the probability that the target moment appears in the wrong videos to close to zero, so  $P(m_*|\mathcal{V}_*,q)$ approximately equals to $P(m_*|v_*,q)$. The details of logits $l^{st}_{*,j}$ are introduced in \cref{sec:localization}.

Combine \cref{equ:two_tasks}, \cref{equ:prob_topk} and \cref{equ:prob_localization}, the probability $P(m_*|v_*,q)$ can be computed by:
\begin{equation}
\begin{split}
\label{equ:prob_final}
&P(m_*|v_*,q) \approx \\ 
&\frac{{\rm exp}(S^R_*)}{\sum_{j=1}^{K} {\rm exp}(S^R_j)}  \frac{{\rm exp}(l^{st}_{*,j})}{ \sum \limits_{a=1} \limits^{K} \sum \limits_{b=1} \limits^{|v_i|} {\rm exp}(l^{st}_{a,b}) }  \frac{{\rm exp}(l^{ed}_{*,k})}{\sum \limits_{a=1} \limits^{K}  \sum \limits_{b=1} \limits^{|v_i|} {\rm exp}(l^{ed}_{a,b}) },
\end{split}
\end{equation}
where the denominator is the same for all candidate moments from K videos, so we can simplify this probability to a new scoring function:
\begin{equation}
\label{equ:score_func}
S_* = S^R_* +  l_{*,j}^{st} + l_{*,k}^{ed},
\end{equation}
where $l_{*,j}^{st} + l_{*,k}^{ed} = S^L_{*, ij}$ represents moment localization score. This scoring function is simpler than \cref{equ:score_pre} and without hyper-parameter $\alpha$ which may greatly increase the weight of the top-ranked video retrieval score.

In inference, we use scoring function in \cref{equ:score_func} to rank all moments in multiple retrieved videos.

\subsection{Model}
We propose a two-stage MINUTE model consisting of a late-fusion video retriever and an early-fusion moment localizer.  

\subsubsection{Video Retriever}
\label{sec:retrieval}
\begin{figure}
    \begin{center}
    \includegraphics[width=0.5\textwidth]{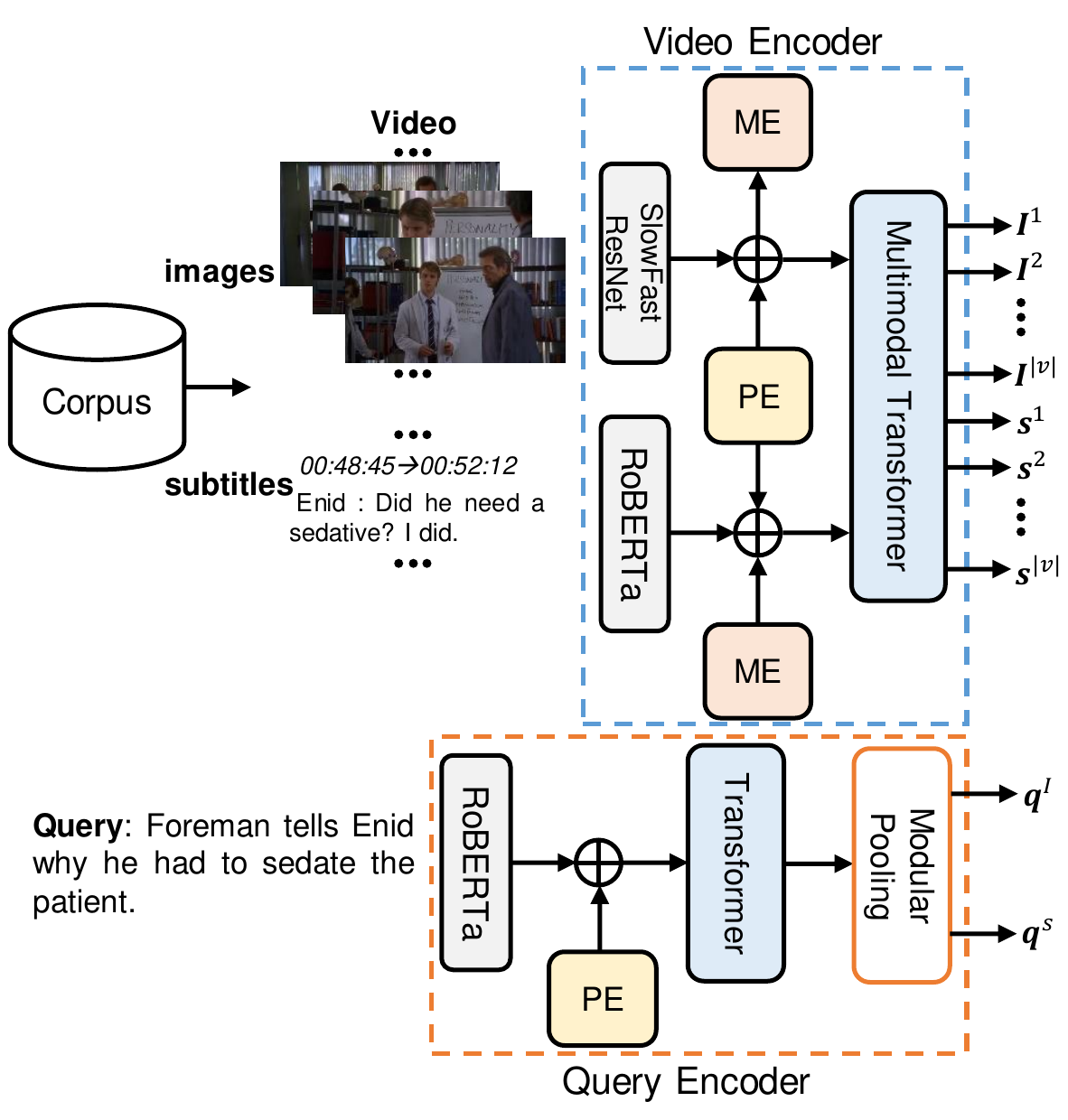}
    \end{center}
    \caption{Video retriever consists of two encoders, video encoder and query encoder. 'ME' and 'PE' represent modality embedding and positional embedding, respectively. }
    \label{fig:retriever}
\end{figure}

The goal of video retriever is to select a small subset $\mathcal{V}^*$ from the corpus $\mathcal{V}$ given the query $q$, where videos in the subset may contain the target moment. 
The retriever of the proposed model is a late-fusion model that contains two encoders, a query encoder and a video encoder, as shown in \cref{fig:retriever}. The late-fusion architecture ensures retrieval efficiency if we index the representations of videos in advance. 

\noindent \textbf{Video Encoder} The video encoder encodes frames in the $i$-th video to frame representations $\bm{v}_i = \{\bm{f}_i^1, ...,\bm{f}_i^{|v_i|}\}$, where  the $j$-th frame $\bm{f}_i^j$ contains image representation $\bm{I}_i^j$ and subtitle representation $\bm{s}_i^j$. We first use RoBERTa~\cite{liu2019roberta} to extract sentence features of subtitle and use SlowFast~\cite{feichtenhofer2019slowfast} and ResNet~\cite{he2016deep} to extract image  features. Then we feed subtitle features and image features to a one-layer multi-modal Transformer that simultaneously captures intra-modal and inter-modal dependencies to output each image representation $\bm{I}_i^j$ and subtitle representation $\bm{s}_i^j$. 

\begin{figure}
    \begin{center}
    \includegraphics[width=0.48\textwidth]{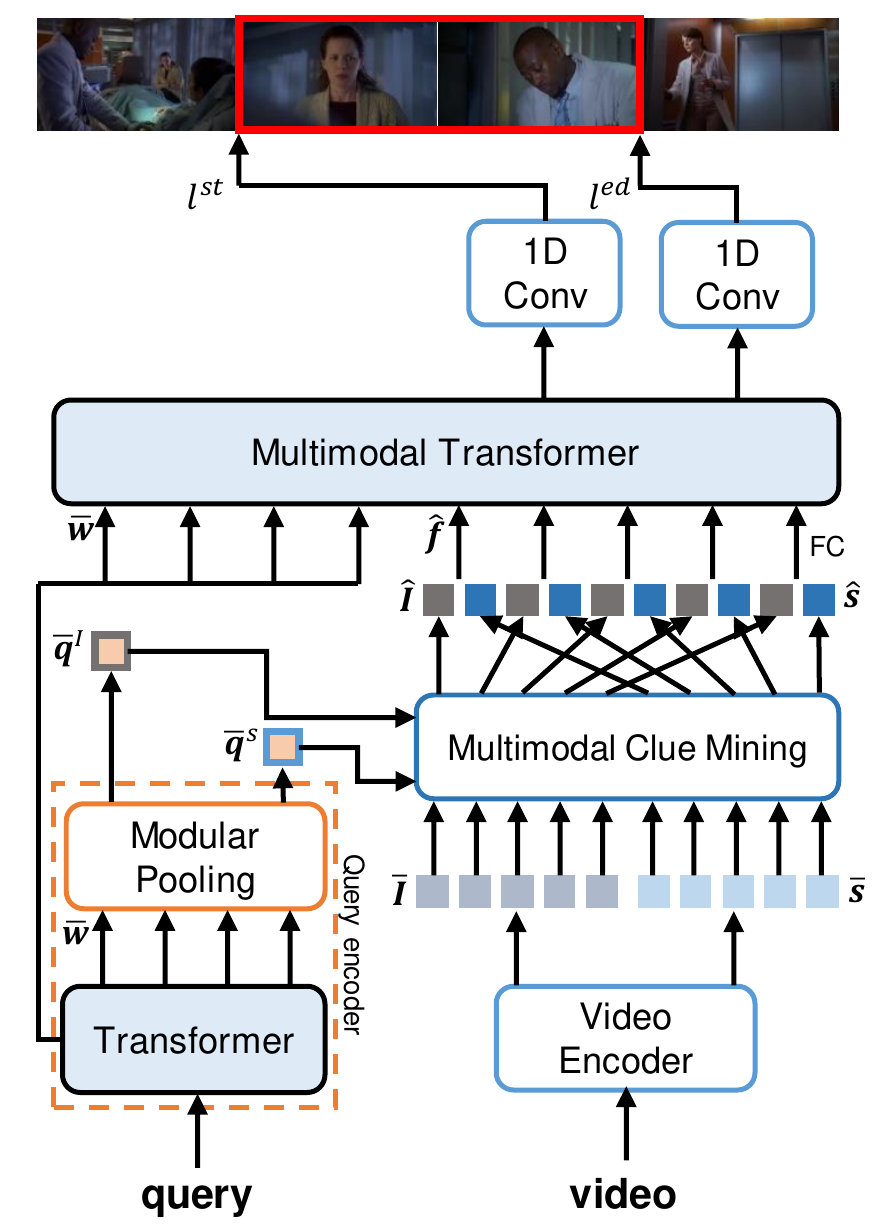}
    \end{center}
    \caption{Moment localizer contains two components, multimodal clue mining and multimodal Transformer. For brevity, we omit the subscripts of the representations.}
    \label{fig:localizer}
\end{figure}

\noindent \textbf{Query Encoder} The query encoder convert query $q=\{w^1,w^2,...,w^{|q|}\}$ to query representation $\bm{q}$. We first use RoBERTa to extract the feature $\bm{w}^j$ of each word in the query. A one-layer Transformer is used to capture the contextual representation of each word.  We generate two query representations for query-image similarity score and query-subtitle similarity score, denoted as $\bm{q}_I$ and $\bm{q}_s$. We adopt a modular pooling mechanism~\cite{lei2020tvr} to convert the sequence representations to the two vectors:
\begin{equation} 
o^i = \bm{W}_c\bm{w}^i   , \\
\ \alpha^i = \frac{{\rm exp}(o^i)}{\sum \limits_{j=1}^{|q|} {\rm exp}(o^j)}   , \\
\ \bm{q}_c =  \sum \limits_{i=1}^{|q|} \alpha^i \bm{w}^i  ,
\end{equation}
where $\bm{W}_c$ is learnable parameters, $c \in \{I, s\}$. The modular mechanism can be regarded as a learnable pooling and is also used in previous works\cite{lei2020tvr, li2020hero, zhang2021video}.

We also use the retrieval head of HERO~\cite{li2020hero} as retriever for a fair comparison with CONQUER~\cite{hou2021conquer}. The original HERO uses margin-based loss~\cite{faghri2017vse++} to train video retrieval whose retrieval score only represents cosine similarity between query and videos, so we re-train HERO in the same way as training the proposed retriever to model the probability $P(v_* | q, \mathcal{V})$ in~\cref{equ:prob_topk}. We use \textbf{simple retriever} to denote the proposed retriever and \textbf{HERO retriever} to denote the retriever based on HERO.

\subsubsection{Moment Localizer}
\label{sec:localization}
Moment localizer shown in \cref{fig:localizer} uses the query to localize the target moment $m_*$ in the top-K retrieved videos $\mathcal{V}_*$. The proposed localizer is based on early-fusion architecture to explore deeper interactions between  query and video. 
Because the retrieved videos are narrowed down to a small range, the amount of computations is acceptable.

The localizer first uses query encoder to get token representations $\{\bm{\bar{w}}^1,...,\bm{\bar{w}}^{|q|}\}$ and video encoder to get video representation $\bm{\bar{v}}_i = \{\bm{\bar{f}}_i^1, ...,\bm{\bar{f}}_i^{|v_i|}\}$, where $\bm{\bar{f}}_i^j$ contain an image representation and a subtitle representation $(\bm{\bar{I}}_i^j, \bm{\bar{s}}_i^j)$. Video encoder and query encoder in localizer are same with those in retriever but do not share parameters. 

Our proposed localizer consists of two components: multimodal clue mining and multi-modal Transformer.

\noindent \textbf{Multimodal Clue Mining (MCM)} solves late key content problem by discovering important content from multiple modalities of video to help moment localization. MCM first uses query to measure the importance of each image and subtitle in video, then assigns weights to these elements from different modalities according to importance. 

Specially, we leverage modular pooling to obtain query representations $\bm{\bar{q}}^I$ and $\bm{\bar{q}}^s$ to measure image importance and subtitle importance respectively. The importance is computed by:
\begin{equation} 
\bm{p}^j_c = (\bm{\bar{W}}_c\bm{\bar{c}}^j)  \odot  \bm{\bar{q}}^c  , c \in \{I ,s\}  , 
\end{equation}
where $\bm{\bar{W}}_c$ is learnable parameters, and $\bm{p}^j_c$ is the importance of $j$-th image or subtitle. Then we use the importance to weight the image and subtitle representations:
\begin{equation} 
\bm{\hat{c}}^j = {\rm norm}(\bm{p}^j_c) \odot \bm{\bar{c}}^j    , c \in \{I ,s\}  ,
\end{equation}
where $\bm{\hat{c}}^j$ is weighted image representation or subtitle representation and ${\rm norm}$ is L2-normalization which makes the model converge better.

MCM can be seen as an amplifier that allows localizer to focus on important content which we call clues from multiple modalities.

We fuse the weighted representations $\bm{\hat{I}}^j$ and $\bm{\hat{s}}^j$ in a frame by a fully-connect layer:
\begin{equation} 
\bm{\hat{f}}^j =  {\rm FC}([\bm{\hat{I}}^j;\bm{\hat{s}}^j])    , 
\end{equation}
where $[;]$ is concatenation and $\bar{\bm{f}}^j$ is the fused representation the j-th frame. The fused video representation is $\bm{\hat{v_i}} = \{\bm{\hat{f}}^1_i, ..., \bm{\hat{f}}^{|v_i|}_i \}$ and are fed to a multimodal Transformer together with query token representations.

\noindent \textbf{Multimodal Transformer (MMT)} 
We use  a three-layer multi-modal Transformer to make deep interactions between fused video representation and token representations.  In addition, two 1D-convolution layers are leveraged to capture dependencies between adjacent frames and output logits  $l^{st}_{i,j}$, $l^{ed}_{i,k}$  of the start and end times of the target moment.

\subsection{Training}
We first train retriever by text-video pairs, then use the trained retriever to sample negative videos as hard negatives to train localizer.

\noindent \textbf{Training retriever} To maximize the log-likelihood of probability ${\rm log} P(v_* | q, \mathcal{V})$ in \cref{equ:prob_retrieval}, we adopt InfoNCE~\cite{van2018representation} loss with in-batch negative sampling to train retriever. Specially, let $d = \{ (v_1, q_1), ..., (v_b, q_b) \}$ denote training data in a batch, where $b$ is batch size. Each pair $(v_i, q_i)$ in $d$ has $b - 1$ negative samples for query-to-video loss or video-to-query loss, such $(v_z,q_i)_{z \ne i}$ and  $(v_i,q_z)_{z \ne i}$:
\begin{equation}
\begin{aligned}
\mathcal{L}^v =  -log \frac{{\rm exp}(S^R_{i,i})}{\sum\limits_{z = 1}^{b} {\rm exp}(S^R_{z,i})}, 
\mathcal{L}^q =  -log \frac{{\rm exp}(S^R_{i,i})}{\sum\limits_{z = 1}^{b} {\rm exp}(S^R_{i,z})},
\end{aligned}
\end{equation}
where $\mathcal{L}^v$ and $\mathcal{L}^q$ are query-to-video loss and video-to-query loss,  respectively. We use the sum of the two losses to train retriever.

\noindent \textbf{Training localizer}
We use the well-trained retriever to retrieve top-ranked videos from training data and sample n videos as hard negatives to train the localizer with Shared-Norm technique.
\begin{equation}
\mathcal{L}^{st} = -log   \frac{{\rm exp}(l^{st}_{+,j})}{\sum \limits_{a=1} \limits^{n+1} \sum \limits_{b=1} \limits^{|v_b|} {\rm exp}(l^{st}_{a,b}) }  , 
\mathcal{L}^{ed} = -log   \frac{{\rm exp}(l^{ed}_{+,k})}{\sum \limits_{a=1} \limits^{n+1} \sum \limits_{b=1} \limits^{|v_b|} {\rm exp}(l^{ed}_{a,b}) } , 
\end{equation}
The sum of $\mathcal{L}^{st}$ and $\mathcal{L}^{ed}$ are used to train localizer. 


\begin{table}
\caption{Comparisons of VCMR results(IoU=0.7) with baselines on TVR validation set and testing set.'SR' denotes simple retriever, and 'HR' denotes HERO retriever.}
\centering
\scalebox{0.85}{
\begin{tabular}{lllllll}
\hline
\multirow{2}{*}{Model} & \multicolumn{3}{c}{Validation} & \multicolumn{3}{c}{Testing} \\
                                    & R1     & R10    & R100   & R1     & R10    & R100 \\
\hline
XML                & 2.62    & 9.05    & 22.47   & 3.32    & 13.41   & 30.52\\
ReLoCLNet     & 4.15    & 14.06   & 32.42   & -       & -       & -\\
HAMMER    & 5.13    & 11.38   & 16.71   & -       & -       & -\\
HERO              & 5.13    & 16.26   & 24.55   & 6.21    & 19.34   & 36.66\\
CONQUER            & 7.76    & 22.49   & 35.17   & 9.24    & 28.67   & 41.98\\
\hline
MINUTE(SR)     & 8.17   & 23.38   & 37.93  & 9.59  & 28.96 & 45.23 \\
MINUTE(HR)    & \textbf{10.70}   & \textbf{29.37}   & \textbf{45.09}  & \textbf{12.60}  & \textbf{33.72} & \textbf{50.23} \\
\hline
\end{tabular}
}
\label{tab:vcmr_tvr}
\end{table}

\begin{table}
\caption{Comparisons of VCMR results with baselines on DiDeMo  testing set.}
\centering
\scalebox{0.85}{
\begin{tabular}{lllllll}
\hline
\multicolumn{1}{c}{\multirow{2}{*}{Model}} & \multicolumn{3}{c}{IoU=0.5} & \multicolumn{3}{c}{IoU=0.7}             \\
\multicolumn{1}{c}{}        & R1      & R5      & R10     & R1      & R5   & R10   \\
\hline 
XML        & 2.36    & -       & 10.42   & 1.59   & -    & 6.77  \\
HERO      & 3.37    & 8.97    & 13.26   & 2.76   & 7.73 & 11.78 \\
CONQUER   & 3.31    & 9.27    & 13.99   & 2.79   & \textbf{8.04} & 11.90 \\
\hline
MINUTE(HR)              & \textbf{3.44}    & \textbf{9.62}    & \textbf{14.62}   & \textbf{2.81}   & 7.89 & \textbf{12.03} \\
\hline
\end{tabular}
}
\label{tab:vcmr_didemo}
\end{table}

\section{Experiment}
We first introduce datasets and metrics. Then we describe implementation details. After that, we introduce experimental results comparison with baselines. Then we illustrate ablation studies of the proposed model. Finally, we present the case study.

\subsection{Datasets}
\label{sec:dataset}
\noindent \textbf{TVR}\cite{lei2020tvr} is built on TV Shows whose videos consist of images and subtitles. TVR contains 17435, 2179, and 1089 videos on the training, validation, and testing sets.  
The average length of the videos is 76.2 seconds, while the average length of the moments is 9.1 secs.

\noindent \textbf{DiDeMo}\cite{anne2017localizing} is a dataset whose videos are from the real world, with only images and no subtitles in the video. DiDeMo contains 8395, 1065, and 1004 training, validation, and testing videos, respectively. 
The average duration of videos and moments is 54 secs and 6.5 secs, respectively. 

\subsection{Evaluation Metrics }

We follow the metrics in \cite{lei2020tvr} as evaluation metrics of experiments. For VCMR task, the evaluation metric is \textbf{R@$K$, IoU=$p$} that represents the percentage that at least one predicted moments whose Intersection over Union(IoU) with the ground truth exceed $p$ in the top-K retrieved moments. The two sub-tasks are also evaluated. The metric of SVMR task is the same as that of VR task, but the evaluation is conducted in only ground truth video for each query. As for VR task, the metric is  \textbf{R@$K$} which denotes the percentage that correct video is in the top-K ranked videos.

\begin{table}
\caption{Comparisons of VR results with baselines on TVR validation set.}
\centering
\scalebox{0.85}{
\begin{tabular}{lllll}
\hline
Model         & R@1   & R@5   & R@10  & R@100 \\
\hline
XML           & 16.54 & 38.11 & 50.41 & 88.22 \\
ReLoCLNet     & 22.13 & 45.85 & 57.25 & 90.21 \\
HERO          & 29.01 & 52.82 & 63.07 & 89.91 \\
\hline
SR            & 23.12 & 46.86 & 57.83 & 90.22 \\
HR & \textbf{32.88}  & \textbf{55.62} & \textbf{65.35} & \textbf{91.26} \\
\hline
\end{tabular}
}
\label{tab:vr}
\end{table}

\begin{table}
\caption{Comparisons of SVMR results with baselines on TVR Validation set.}
\centering
\scalebox{0.85}{
\begin{tabular}{lllllll}
\hline
\multirow{2}{*}{Model} & \multicolumn{3}{c}{IoU=0.5} & \multicolumn{3}{c}{IoU=0.7} \\
                       & R1     & R10    & R100   & R1     & R10    & R100   \\
\hline
XML                    & 31.43   & -       & -       & 13.89    & -   & -   \\
ReLoCLNet              & 31.88   & -       & -   & 15.04    & -   & -  \\
HERO                   & 32.22   & 60.08   & 80.66   & 15.30    & 40.84   & 63.45  \\
CONQUER                & 43.63   & -       & -       & 22.84    & -  & -   \\
\hline
MINUTE(SR)                     & 44.49   & 78.62    & 93.57   & 23.98    & 61.30 & 80.13 \\
MINUTE(HR)        & \textbf{44.74}   & \textbf{78.90}   & \textbf{93.80}   & \textbf{24.08}    & \textbf{62.10}   & \textbf{80.45}  \\
\hline
\end{tabular}
}
\label{tab:svmr}
\end{table}

\subsection{Implementation Details}

\noindent \textbf{Training} We train simple retriever for 100 epochs with the batch size 256. As for localizer, we sample 4 and 2 negative videos for each query from top-100 ranked videos on TVR and DiDeMo respectively, and train it for 10 epochs with the batch size 32. Both simple retriever and localizer are trained by AdamW with the learning rate 0.0001 and the weight decay of 0.01 in a single 3090 GPU. For HERO retriever, we retrain it with InfoNCE loss in 8 3090 GPUs with the same setting as the original HERO~\cite{li2020hero}.

\noindent \textbf{Inference} The localizer localizes the target moment in the top-10 retrieved videos. The length of predicted moments is limited to $[1, 24]$ and $[1, 7]$ for TVR and DeDiMo, respectively. We use non-maximum suppression(NMS) with the IoU 0.7 to post-process the predicted moments.

\subsection{Comparison with Baselines}
We compare the proposed model with baselines on VCMR task including four one-stage models XML~\cite{lei2020tvr}, ReLoCLNet~\cite{zhang2021video}, HAMMER~\cite{zhang2020hierarchical}, HERO~\cite{li2020hero} and a two-stage model CONQUER~\cite{hou2021conquer}.

\noindent \textbf{TVR} As shown in~\cref{tab:vcmr_tvr}, the proposed models outperform all baseline methods. Compared with the best previous method CONQUER who also uses HERO to address the VR task, our proposed model with HERO retriever achieves 36\% improvement at R@1 on the testing set. We also report the results on two sub-task in~\cref{tab:vr} and~\cref{tab:svmr}. For  VR, HERO retriever trained by InfoNCE loss has better retrieval accuracy than the original HERO. For SVMR, our proposed models also achieve the best results. It is worth noting that the proposed model with simple retriever outperforms CONQUER on VCMR even though the performance of VR(R@1 23.12) is much worse than that in CONQUER(R@1 29.01). This is because moment prediction bias limits the performance of CONQUER.

\noindent \textbf{DiDeMo} We report the VCMR results on DiDeMo testing set in ~\cref{tab:vcmr_didemo}. The performance of the proposed model is still better than others. All the methods perform worse than the results on TVR because the DiDeMo dataset is designed for temporal language grounding, so the difficulty of retrieving video is not considered. The query of DiDeMo is not as specific as that of TVR, such as "a girl is playing ball", making it hard to retrieve the correct video.

\begin{figure}[!tbp]
    \begin{center}
    \includegraphics[width=0.28\textwidth]{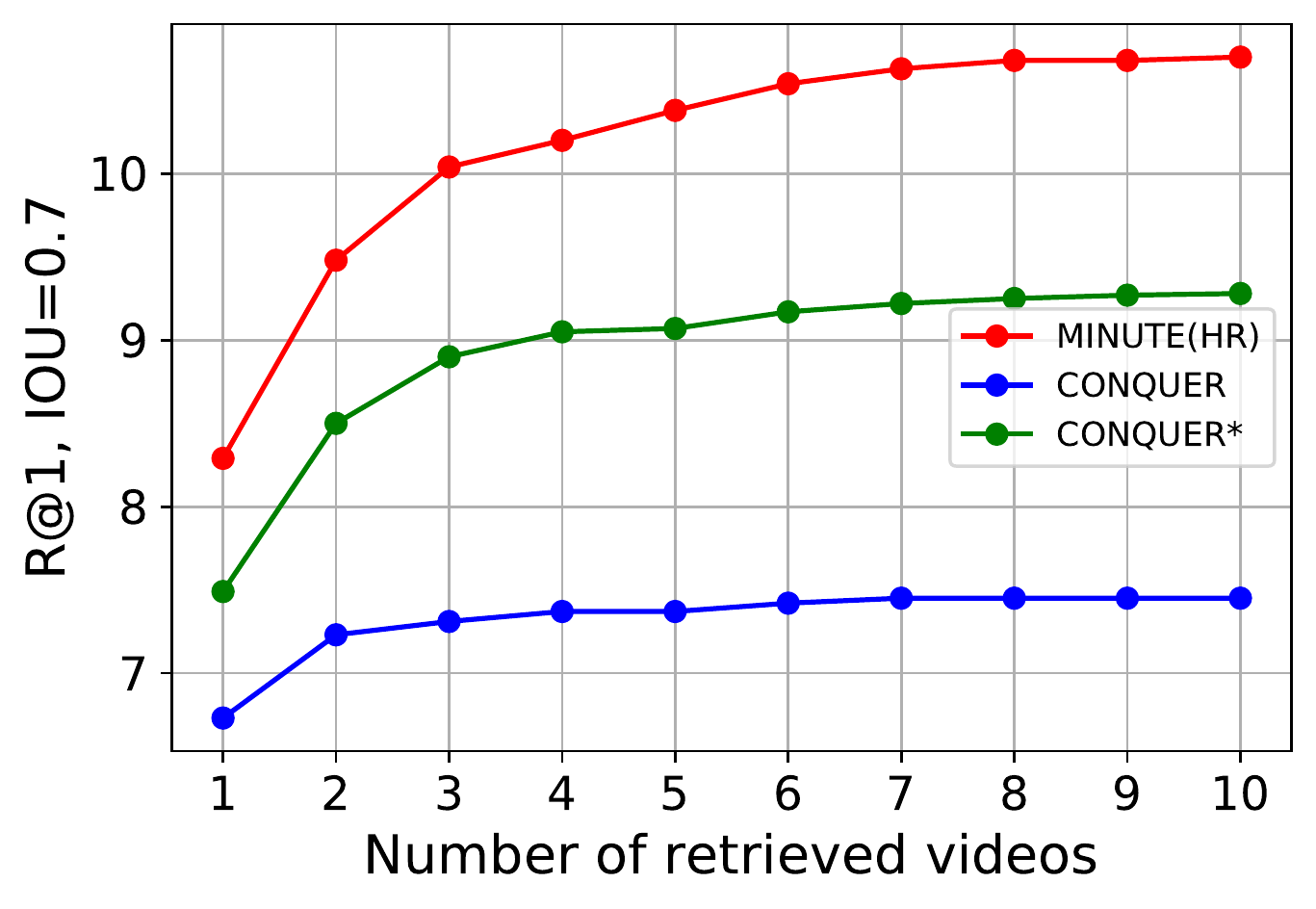}
    \end{center}
    \caption{The performances of VCMR of our model and COUQUER under different numbers of the retrieved videos, where 'CONQUER*' denotes CONQUER with our retriever and scoring function.}
    \label{fig:k_nunm}
\end{figure}

\begin{table}
\caption{Performances of VCMR and SVMR (R@1, IOU=0.5,0.7) when remove two componets in localizer. MCM denotes multimodal clue mining, and MMT represents multimodal Transformer.}
\centering
\scalebox{0.85}{
\begin{tabular}{lllll}
\hline
\multicolumn{1}{c}{\multirow{2}{*}{Model}} & \multicolumn{2}{c}{VCMR} & \multicolumn{2}{c}{SVMR} \\
\multicolumn{1}{c}{}                       & 0.5         & 0.7        & 0.5         &   0.7    \\
\hline
MINUTE(HR)                             & \textbf{19.22}       & \textbf{10.70}     & \textbf{44.74}       & \textbf{24.08}      \\
w/o MCM                                    & 18.21       & 10.17      & 43.41       & 23.46      \\
w/o MMT                                    & 16.71       & 8.66       & 40.5        & 20.97      \\
\hline
\end{tabular}}
\label{tab:ab_localizer}
\end{table}

\subsection{Moment Prediction Bias}
As shown in \cref{fig:k_nunm}, when the number of retrieved videos increases, the performance of our model improves, but the CONQUER does not change much, which indicates that moment prediction bias limits its performance. This bias is from the inconsistency of Shared-Norm in training and inference. Our prediction based on the scoring function in \cref{equ:score_func} addresses this prediction bias by ranking moments in multiple retrieved videos in inference. When we replace CONQUER's retriever and scoring function with ours, CONQUER* in \cref{fig:k_nunm} can also improve moment prediction bias, showing the proposed model's effectiveness.

\subsection{Multimodal Clue Mining}
We perform ablation studies on the effectiveness of two components of localizer in~\cref{tab:ab_localizer}.  
When removing MCM, the accuracy drops, which shows that discovering key content from images and subtitles as clue is helpful for moment localization. When we only use MCM, the accuracy drops a lot, indicating that using clues is not enough, fine-grained cross-modal interactions are also needed.

\subsection{Case Study}
\begin{figure}
    \begin{center}
    \includegraphics[width=0.5\textwidth]{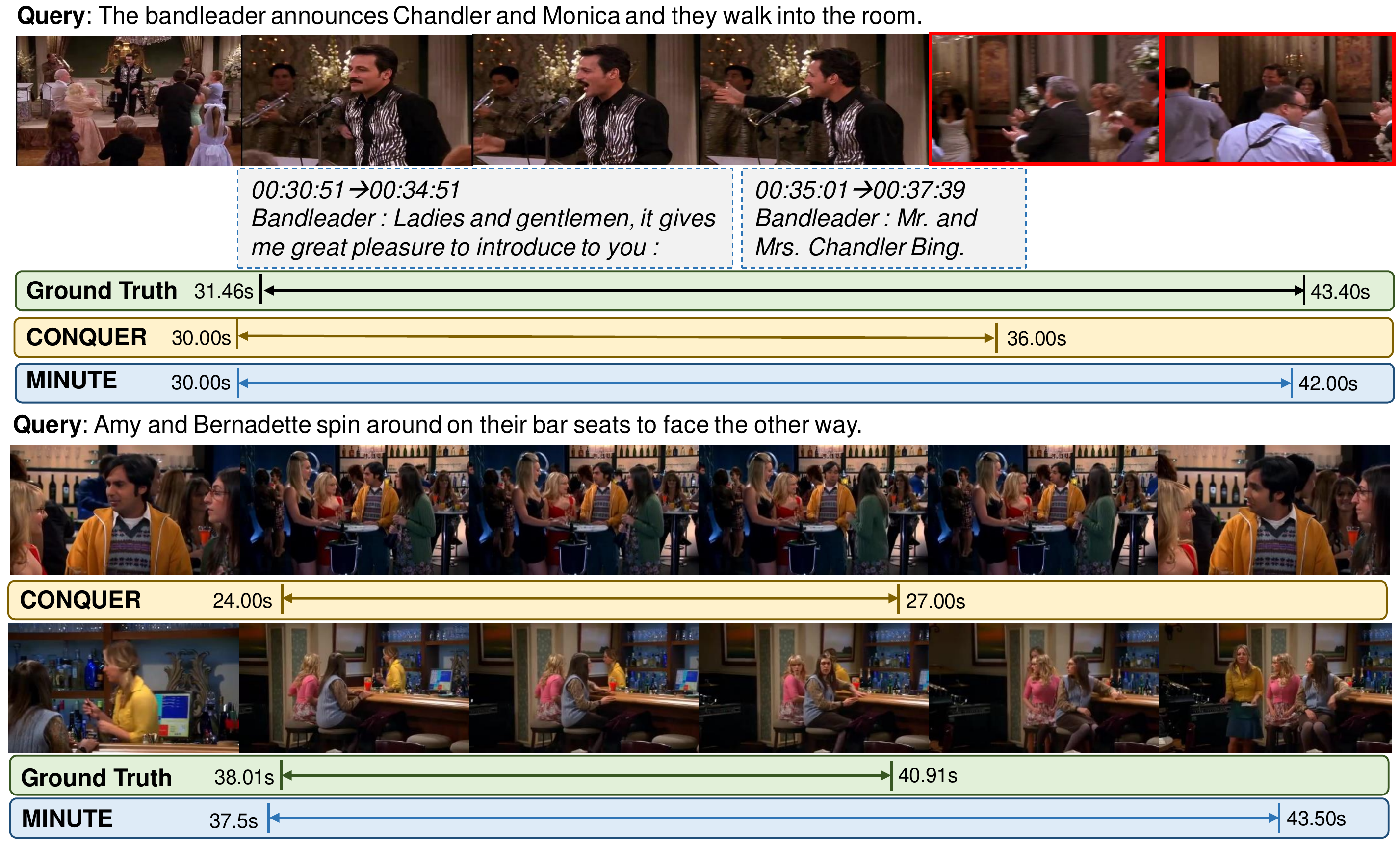}
    \end{center}
    \caption{Two cases on TVR from the proposed model and CONQUER.}
    \label{fig:case}
\end{figure}

We show two cases of VCMR in~\cref{fig:case}. In the first case, two models retrieve the correct video first, the moment predicted by the proposed model is closer to the ground truth. The proposed model captures key images related to  "they walk into the room" to help localize the moment, indicating the effectiveness of MCM in our model. In the second case, both models rank the wrong video first because the scenario in this video is similar to that in the correct video. CONQUER fails to predict correct moment from correct video, for it places too much emphasis on top-ranked videos. Our proposed model can predict correct moment, which verifies that our prediction improves moment prediction bias.

\section{Conclusion}
In this paper, we  propose a model \textbf{M}ult\textbf{I}-video ra\textbf{N}king with m\textbf{U}l\textbf{T}imodal clu\textbf{E}~(MINUTE) improving two problems of two-stage method on video corpus moment retrieval task, moment prediction bias and latent key content. We first analyze the reason for moment prediction bias that inconsistency of Shared-Norm in training and inference, then we adopt Shared-Norm in inference and rank moments in multiple videos based on our derived scoring function to improve moment prediction bias. As for latent key content, we propose a multimodal clue mining component to discover important content from two modalities of video as clue for better moment localization. Extensive experiments on two datasets TVR and DiDeMo show that our proposed model improves two problems and achieves a new state-of-the-art of video corpus moment retrieval task.


{\small
\bibliographystyle{ieee_fullname}
\bibliography{egbib}
}

\end{document}